\documentclass[twocolumn]{article}

\input bangfont

\usepackage{multirow}
\usepackage{graphicx}
\usepackage{pdflscape}
\usepackage{url}
\usepackage{times}
\usepackage{amsmath}

\usepackage{fullpage}
\usepackage{balance}
\usepackage{appendix}

\title{Evolution of the Modern Phase of Written Bangla:\\A Statistical Study}
\author{
	\begin{tabular}{cc}
		Paheli Bhattacharya & Arnab Bhattacharya \\
		\url{pahelibhattacharya@gmail.com} & \url{arnabb@iitk.ac.in} \\
		Govt. College of Engineering
			& Dept. of Computer Science \\
		 and Textile Technology, & and Engineering, \\
		Serampore, Hooghly, & Indian Institute of Technology, Kanpur, \\
		India. & India.
	\end{tabular}
}

\date{}

\newcommand{\comment}[1]{}
\newcommand{\scfm}[1]{\ensuremath{\text{\sc{e}-}#1}}
\newcommand{\tabcaption}[1]{\vspace*{0mm}\caption{#1}\vspace*{0mm}}

\begin{document}

\maketitle

\begin{abstract}
	Active languages such as Bangla (or Bengali) evolve over time due to a
	variety of social, cultural, economic, and political issues.  In this paper,
	we analyze the change in the written form of the modern phase of Bangla
	quantitatively in terms of character-level, syllable-level, morpheme-level
	and word-level features.  We collect three different types of
	corpora---classical, newspapers and blogs---and test whether the differences
	in their features are statistically significant.  Results suggest that there
	are significant changes in the length of a word when measured in terms of
	characters, but there is not much difference in usage of different
	characters, syllables and morphemes in a word or of different words in a
	sentence.  To the best of our knowledge, this is the first work on Bangla of
	this kind.
\end{abstract}

\section{Introduction}
\label{sec:intro}

Bangla (or Bengali) is one of the most widely spoken languages.  It belongs to
the Indo-European family of languages and is believed to have been derived from
Prakrit in around 650\,CE.  The history of Bangla is divided into three phases:
Old Bangla (till 1350\,CE), Medieval Bangla (1350-1800\,CE) and Modern Bangla
(1800\,CE-).\footnote{The history and genesis of the language can be found in
\url{http://www.bpedia.org/B_0137.php}.}

Since its inception, Bangla, like any other active language, has undergone a lot
of changes due to a variety of social, cultural, economic and political causes.
The changes happen mostly in vocabulary and pronunciation, one of the big
catalysts for which is the adoption of words of foreign origin either directly
or indirectly into the language.  For example, in Bangla, there is no word that
depicts the concept ``football'' directly, and consequently, the English word
has been adopted verbatim and has become part of the language now.  Similarly,
the English word ``box'' has been incorporated in Bangla as \bngx bak/s \rm
(bAksa\footnote{We have used the ITRANS transliteration mechanism to specify the
words in Bangla font (\url{http://www.aczoom.com/itrans/}).  The rules for
Bangla are available at
\url{http://www.aczoom.com/itrans/html/beng/node4.html}.}) by suitably modifying
its pronunciation.

A particularly remarkable source of variety in Bangla is the two clearly
distinct forms of written prose in the modern phase -- \emph{Sadhu Bhasha}
(chaste language) and \emph{Chalit Bhasha} (colloquial language).  The chaste
language was used earlier (by the likes of Bankimchandra Chattopadhyay,
Rabindranath Tagore, Saratchandra Chattopadhyay and others) and has been now
replaced in almost all communications in Bangla by the colloquial version.  The
most notable change has happened in the form of verbs and pronouns which has
become shorter and can be more easily pronounced.  For example, the verb \bngx
kiryaich \rm (kariAChi) has become \bngx kerich \rm (karaChi) and the pronoun \bngx
taHaedr \rm (tahadera) has been transformed to \bngx taedr \rm (tader).

With the advancement of digital world, the electronic media have imparted a
large impact on the modern language which is clearly reflected in newspapers,
blogs and social networking forums.  It is extremely rare to find longer words
such as \bngx eJoubnetejadiip/t \rm (JaubanatejodIpta) now and \bngx {oi}ctRinshiithshshii \rm
(chaitaranishIthashashi) than in the classical literature.

However, while all these notions of change are commonly believed to be true, to
the best of our knowledge, there is no work that tests whether these perceptions
about the differences are \emph{statistically significant}.  In this paper, we
precisely aim to fill this gap.  Our main contribution, thus, is to study the
changes in the modern phase of written Bangla in a statistically robust manner.

We collect three different corpora -- one consisting of classical literature,
and the other two that of newspapers and blogs (the details are in
Section~\ref{sec:corpora}).  We then extract different features at the word and
sentence levels and test whether the changes across the corpora are significant
when viewed from a statistical standpoint.

The rest of the paper is organized as follows.  Section~\ref{sec:work} discusses
the related work.  Section~\ref{sec:corpora}, Section~\ref{sec:features} and
Section~\ref{sec:test} describe the corpora, the features and the statistical
testing method respectively.  Section~\ref{sec:results} discusses the results
before Section~\ref{sec:conc} concludes.

\section{Related Work}
\label{sec:work}

\emph{Evolutionary linguistics} or the study of evolution of languages has long
fascinated human beings.  In addition to numerous studies that have been
developed, there are whole conferences---EvoLang, the Evolution of Language
International Conferences (\url{evolang.org})---that are devoted for this.
Among the computational studies for language evolution and change, an
overwhelming majority of the work is focused on European languages
\cite{compevol,langevol,langchange}.

Indian languages, due to the relative paucity of digital resources, had not been
studied deeply.  The recent ease of using Unicode and the surge of excellent
work in the field of natural language processing (NLP) have, however, changed
the situation dramatically.  Sikder analyzed the change in type of words in
Bangla and showed how the frequency of foreign words are increasing, especially
for young people living in urban areas \cite{bangla1}.  Choudhury et al. studied
the change of Bangla verb inflections for the single verb \bngx kr \rm (kara),
which means ``to do'' in English \cite{verb}.  They gave a functional
explanation for the rise in several dialects of Bangla because of phonological
differences while uttering the verb inflections.

To the best of our knowledge, computational studies describing the changes in
the Bangla language, however, have not been undertaken so far.  The evolution of
a language is most visible when changes occur in the discourse.  Since we are
still far off from that for Bangla, in this paper, we study some micro-level
aspects of the written language in terms of characters, syllables, morphemes and
words.

\section{Corpora}
\label{sec:corpora}

For our work, we collected three different corpora:
\begin{enumerate}
	\item \emph{Classical Corpus}: It includes the literary works of 4 eminent
		authors.% as listed in Table~\ref{tab:classical}.
	\item \emph{Newspaper Corpus}: It includes the news articles from 7 leading
		newspapers of both India and Bangladesh.% as listed in
		%Table~\ref{tab:newspaper}.
	%
	\item \emph{Blog Corpus}: It includes blog articles (but not the comments)
		from 11 blogs.% as listed in Table~\ref{tab:blog}.
\end{enumerate}
Appendix~\ref{app:corpora} list the details of the three corpora.
The total number of words in each of the corpus are listed in
Table~\ref{tab:corpora}.

\begin{table}[t]
	\begin{center}
	\begin{tabular}{|c|c|}
	\hline
	{\bf Corpus} & {\bf Number of words} \\ 
	\hline
	\hline
	Classical & 1,58,807 \\
	\hline
	Newspaper & 7,71,989 \\
	\hline
	Blog & 5,18,485 \\
	\hline
	%Merged Corpus & 14,49,281 \\
	%\hline
	\end{tabular}
	\tabcaption{Number of words in the corpora.}
	\label{tab:corpora}
	\end{center}
\end{table}

\section{Features}
\label{sec:features}

\subsection{Character-level Features}

In Bangla, there are two types of characters---vowels and consonants.  The
consonants cannot be pronounced on their own and must always end with the sound
of a vowel.  Vowels, on the other hand, can be pronounced on their own and are
written either as independent letters or as diacritical marks on the consonant
they attach to.  For example, \bngx k:/ \rm (k) is a consonant.  When it is joined
with \bngx Aa \rm (A), it is written as \bngx ka \rm (kA).  Thus, the diacritical
mark for the vowel \bngx Aa \rm (A) is \bngx a \rm (A\footnote{The ITRANS coding
for the diacritical marks remain the same.}).  The vowel \bngx A \rm (a) has an
invisible diacritical mark.  Its only effect is to remove the \ \bngx :/ \rm (the
consonant-ending marker) from the consonant it attaches to.  Thus, \bngx k:/+A \rm
(k + a) is written as \bngx k \rm (ka).

We distinguish the diacritical mark of a vowel from the vowel itself as the
latter can stand on its own.  For example, the correct parsing of \bngx khushiiet \rm
(khushIte) is \bngx kh:/+u+sh:/+ii+t:/+e \rm (kh+u+sh+I+t+e) and that of \bngx Aaelak \rm
(Aloka) is \bngx Aa+l:/+ea+k:/+$\cdot$ \rm (A+l+o+k+a) where $\cdot$ is used to
represent the invisible diacritical mark of the vowel \bngx A \rm (a).  The four
consonants, \bngx t//, NNG, {h}, NN \rm (t.h, .n, H, .N respectively), are treated
differently in that they do not have the consonant-ending marker \ \bngx :/ \rm.
Thus, \bngx baNNGla \rm (bA.nlA) is parsed as \bngx b:/+a+NNG+l:/+a \rm (b+A+.n+l+A).

Conjunct characters where two (or three) consonants are joined together are
parsed differently.  There is no vowel at the end of the first (respectively,
the first two) and only the last consonant has a vowel ending written as a
diacritical mark.  Hence, the correct parsing of \bngx sn/tRs/t \rm (santrasta) is
\bngx s:/+$\cdot$+n:/+t:/+r:/+$\cdot$+s:/+t:/+$\cdot$ \rm (s+a+n+t+r+a+s+t+a).

\subsubsection{Character frequencies}
We count the frequencies of all the characters---consonants, vowels and
diacritical marks---using the parsing system discussed above for the three
corpora.  The number of distinct characters is 61 that includes the 39
consonants (the consonant \bngx b \rm (b) is counted only once), the 11 vowels
(the vowel \bngx 9 \rm (no ITRANS code) is not used any more) and the
corresponding 11 diacritical marks.

We also count the frequencies of bi-gram and tri-gram characters.  For example,
the bi-grams in the word \bngx baNNGla \rm (bA.nlA) are \bngx ba \rm (bA), \bngx  aNNG \rm
(A.n), \bngx  NNGl:/ \rm (.nl) and \bngx la \rm (lA).  The tri-grams are extracted
similarly.

We arrange the uni-gram characters (and bi-grams and tri-grams) in descending
order of their frequencies.  When comparing corpus $C_1$ with $C_2$, we consider
the top-$50$ entries from the sorted list of $C_1$ and find their frequencies in
$C_2$.  Thus, the comparison of $C_1$ with $C_2$ differs from that of $C_2$ with
$C_1$ as, in the later case, the frequencies of the top-$50$ entries of $C_2$
are considered.  The frequencies from the two corpora form the two
non-parametric distributions between which the changes are statistically tested.
Instead of using the raw counts as frequencies, we compute the relative ratios
by dividing by the total number of characters in the corpus; this makes two
corpora of differing sizes comparable.

\subsubsection{Character-based word length}
For each corpus, we produce a count of words that have a particular length in
terms of characters.  Thus, if there are $300$ words of length $4$, the
frequency corresponding to $4$ in the non-parametric distribution is $300$.  The
distribution consists of all the word lengths and their frequencies.  The
comparison of corpus $C_1$ with $C_2$ is symmetric for this feature.

\subsection{Morpheme-level Features}

A \emph{morpheme} is the smallest meaning-bearing unit in a language.  A
morpheme may not be able to stand on its own, although a word necessarily does.
Every word is composed of a root word (sometimes called a \emph{lexeme}) and
possibly one or more morphemes.

To extract morphemes, we used the unsupervised program
Undivide++\footnote{Available from
\url{http://www.hlt.utdallas.edu/~sajib/Morphology-Software-Distribution.html}},
which is based on the work by Dasgupta et al. \cite{undivide}.  Unfortunately,
the program have many parameters, and even after repeated tuning and discussion
with the authors, we could not replicate the accuracies as reported in
\cite{undivide} on our corpora for the 4110-word test-set provided by them.
Although the program is only about 50\% accurate on average, we still use it to
extract all the morphemes from the words in the corpora.
(Appendix~\ref{app:morpheme} reports the performance on the metrics as proposed
in \cite{morphologybangla}.)

\subsubsection{Morpheme frequencies}
For every morpheme, we get a count of words that have it.  Similar to the
character frequencies, we then extract the top-$50$ (normalized) frequencies
from each corpus.

\subsubsection{Morpheme-based word length}
Using the program Undivide++, every word is segmented into a list of prefix(es),
root word and suffix(es).  The ``length'' of the word is then counted as the
number of such segments.  For example, if \bngx pRedshiTek \rm (pradeshaTike) is
segmented into the prefix \bngx pR \rm (pra), the root \bngx edsh \rm (desha) and
the two suffixes \bngx iT \rm (Ti) and \bngx ek \rm (ke), its length is counted as
4.

\subsection{Syllable-level Features}

\emph{Syllables} are the smallest subdivisions uttered while pronouncing a word.
Since syllables are phonetic units, they cannot be extracted completely
correctly without speech analysis.  To bypass the problem, we employ a very
simple and intuitive heuristic which is almost always correct.

We assume that any combination of characters till the next vowel is a syllable.
Thus, each vowel, each consonant with its vowel ending (encoded as a diacritical
mark), and each conjunct character is a separate syllable.

The consonants, \bngx t//, NNG, {h} \rm (t.h, .n, H respectively), are treated as single
syllables since they do not have the consonant-ending marker \ \bngx :/ \rm.
However, \ \bngx   NN \rm (.N) is considered part of the preceding syllable.
Thus, the word \bngx AksMat// \rm (akasmAt.h) has three syllables \bngx A \rm (a),
\bngx k \rm (ka), \bngx sMa \rm (smA) and \bngx t// \rm (t.h) while \bngx bNNadha \rm
(bA.NdhA) has two syllables \bngx bNNa \rm (ba.N) and \bngx dha \rm (dhA). 

\subsubsection{Syllable frequencies}
For every uni-gram syllable (and also bi-grams of syllables), we get a count of
words that have it.  We again consider only the top-$50$ (normalized)
frequencies from each corpus.

\subsubsection{Syllable-based word length}
Similar to characters, the word length is also counted in terms of syllables.
The ``inverted list'', i.e., the number of words having a particular
syllable-length is then used as the feature for that syllable length.

\subsection{Word-level Features}

The words are parsed from the sentences using orthographic word boundaries
(i.e., the white-space characters including ?, !, . and the Bangla character
\bngx . \rm).

\subsubsection{Word frequencies}
The words are for sentences what the characters are for words.  Thus, this
feature is computed in exactly the same way as characters.

\subsubsection{Word-based sentence length}
Similar to word length, the sentence length is counted in terms of number of
words.

\begin{table*}[t]
	\begin{center}
		\begin{tabular}{|c||c|c|c|c|}
			\hline
			\multirow{2}{*}{{\bf Feature type}} & \multicolumn{4}{c|}{\bf Level} \\
			\cline{2-5}
			& {\bf Character} & {\bf Syllable} & {\bf Morpheme} & {\bf Word} \\
			\hline
			\hline
			Uni-gram frequency & yes & yes & yes & yes \\
			\hline
			Bi-gram frequency & yes & yes & no & yes \\
			\hline
			Tri-gram frequency & yes & no & no & no \\
			\hline
			Length of word or sentence & yes & yes & yes & yes \\
			\hline
		\end{tabular}
		\tabcaption{Features used.}
		\label{tab:features}
	\end{center}
\end{table*}

\comment{

\begin{table*}[t]
	\begin{center}
		\begin{tabular}{|c|c|}
			\hline
			{\bf Level} & {\bf Feature} \\
			\hline
			\hline
			\multirow{4}{*}{Character} & Uni-gram frequency \\
			 \cline{2-2}
			 & Bi-gram frequency \\
			 \cline{2-2}
			 & Tri-gram frequency \\
			 \cline{2-2}
			 & Word length \\
			\hline
			\hline
			\multirow{3}{*}{Syllable} & Uni-gram frequency \\
			 \cline{2-2}
			 & Bi-gram frequency \\
			 \cline{2-2}
			 & Word length \\
			\hline
			\hline
			\multirow{2}{*}{Morpheme} & Uni-gram frequency \\
			 \cline{2-2}
			 & Segmentation length \\
			\hline
			\hline
			\multirow{3}{*}{Word} & Uni-gram frequency \\
			 \cline{2-2}
			 & Bi-gram frequency \\
			 \cline{2-2}
			 & Sentence length \\
			\hline
		\end{tabular}
		\tabcaption{Features used.}
		\label{tab:features}
	\end{center}
\end{table*}

}

\section{Statistical Testing}
\label{sec:test}

All the features that are used in this paper are summarized in
Table~\ref{tab:features}.

To test whether the distributions of the various features for the different
corpora are statistically different from each other, we employ the
non-parametric two-sample \emph{Kolmogorov-Smirnov (K-S) test}\footnote{We use
the Octave software to perform the tests.}.  For each pair of corpora, we
perform three tests.  Suppose the corpora are $C_1$ and $C_2$.  The \emph{null
hypothesis} $H_0$ for all the three tests state that the samples observed
empirically for $C_1$ and $C_2$ come from the \emph{same} distribution.

There can be three ways by which the \emph{alternate hypothesis} can vary.  For
the non-equal ($\neq$) test, the alternate hypothesis $H_A^\neg$ states that the
empirical values $x_i^{(1)}$ and $x_i^{(2)}$ for the distributions from $C_1$
and $C_2$ are different, i.e., for every $i$, $x_i^{(1)} \neq x_i^{(2)}$.  For
the greater than ($>$) test, the alternate hypothesis $H_A^>$ states that the
empirical values for $C_1$ are greater than the corresponding values for $C_2$,
i.e., for every $i$, $x_i^{(1)} > x_i^{(2)}$.  The less than ($<$) test is
similar where the alternate hypothesis $H_A^<$ tests whether for every $i$,
$x_i^{(1)} < x_i^{(2)}$.

The K-S test returns a \emph{p-value} that signifies the confidence with which
the null hypothesis can be rejected.  The lower the p-value, the more
statistically significant the result is.  Thus, for the $H_A^\neg$ case, it
means the two distributions are more different.  If the result of a $\neq$ test
is statistically significant at a particular level of significance, then the
result of either the $>$ test or the $<$ test (but not both) must be significant
as well at the same level of significance.

\section{Results}
\label{sec:results}

The differences between the word lengths in terms of number of characters
between the three corpora are found to be statistically
significant\footnote{Unless otherwise mentioned, we consider the level of
significance to be 5\%, i.e., a result is statistically significant when the
p-value of the test is less than or equal to $0.05$.} for the alternate
hypothesis $H_A^\neg$.  (The tables in Appendix~\ref{app:results} list all the
p-values.) More interestingly, the alternate hypothesis $H_A^<$ is found to be
very significant for classical versus blog, classical versus newspaper and blog
versus newspaper comparisons.  This shows that the frequency of words having a
shorter length is less in classical than in blogs which, in turn, is less than
newspapers.  Thus, this shows that longer words were more common in the
classical literature than in newspapers which are more than that in blogs.

Although the classical corpus exhibits longer words in terms of syllables (due
to the $H_A^<$ test), the non-equality test ($H_A^\neg$) is not significant.
This, thus, indicates that the use of conjunct characters were more in classical
literature which led to longer words in terms of characters but not in terms of
syllables.  The blogs and newspapers differ in terms of number of syllables
though.

The differences in number of morphemes is again not significant.  Thus, contrary
to popular perception, words with many suffixes and prefixes are not more
abundant in the classical literature as compared to the current scenario.
Similarly, the number of words per sentence for classical is not statistically
different either.

Frequencies of uni-gram characters, bi-gram characters and uni-gram syllables
are not significantly different across the corpora.  Frequencies of tri-gram
characters, bi-gram syllables, uni-gram words and bi-gram words of classical are
significantly different from both blogs and newspapers for the alternate
hypotheses $H_A^\neg$ and $H_A^<$.  The newspaper and blog corpora show little
statistical difference in frequencies indicating that the current styles of
formal and informal writing are quite alike.

\section{Conclusions}
\label{sec:conc}

In this paper, we provided a model of statistically testing the differences of
writing styles across various phases of a language.  To the best of our
knowledge, this is the first work of its kind in Bangla.  This work has aimed at
building a basic foundation on which more analysis in terms of higher-level
features can be carried out in the future.  Also, bigger corpora will allow
robust and more detailed analyses of the results.

\bibliographystyle{abbrv}
\balance
\bibliography{references2013}

%\pagebreak

\appendix

\section*{Appendices}

\section{Corpora}
\label{app:corpora}

Table~\ref{tab:classical}, Table~\ref{tab:newspaper} and Table~\ref{tab:blog}
list the details of the three corpora.

\begin{table*}[pt]
	\begin{center}
		\begin{tabular}{|c|c|c|}
			\hline
			\textbf{Author} & \textbf{URL} & \textbf{On} \\
			\hline
			\hline
			Rabindranath Tagore & \url{www.rabindra-rachanabali.nltr.org}
				& 14$^\text{th}$ July, 2013 \\
			\hline
			Bankimchandra Chattopadhyay & \url{www.bankim.rachanabali.nltr.org}
				& 14$^\text{th}$ July, 2013 \\
			\hline
			Saratchandra Chattopadhyay & \url{www.sarat-rachanabali.nltr.org}
				& 14$^\text{th}$ July, 2013 \\
			\hline
			Swami Vivekananda & \url{www.dduttamajumder.org/baniorachana}
				& 14$^\text{th}$ July, 2013 \\
			\hline
		\end{tabular}
		\tabcaption{Classical corpus.}
		\label{tab:classical}
	\end{center}
\end{table*}

\begin{table*}[pt]
	\begin{center}
		\begin{tabular}{|c|c|c|c|}
			\hline
			\textbf{Name} & \textbf{Website} & \textbf{From} & \textbf{To} \\
			\hline
			\hline
			Anandabazar Patrika & \url{www.anandabazar.com} &
				1$^\text{st}$ June, 2011 & 12$^\text{th}$ July, 2013 \\
			\hline
			Akhon Samay & \url{www.akhonsamoy.com} &
				14$^\text{th}$ July, 2013 & 14$^\text{th}$ July, 2013 \\
			\hline
			Jana Kantha & \url{www.dailyjanakantha.com} &
				1$^\text{st}$ January, 2010 & 13$^\text{th}$ July, 2013 \\
			\hline
			Inqilab & \url{www.dailyinqilab.com} &
				1$^\text{st}$ June, 2013 & 11$^\text{th}$ July, 2013 \\
			\hline
			Jugantor & \url{www.jugantor.com} &
				1$^\text{st}$ July, 2013 & 12$^\text{th}$ July, 2013 \\
			\hline
			Naya Diganta & \url{www.dailynayadiganta.com} &
				30$^\text{th}$ June, 2013 & 14$^\text{th}$ July, 2013 \\
			\hline
			Pratham Alo & \url{www.prothom-alo.com} &
				1$^\text{st}$ January, 2007 & 10$^\text{th}$ July, 2013 \\
			\hline
		\end{tabular}
		\tabcaption{Newspaper corpus.}
		\label{tab:newspaper}
	\end{center}
\end{table*}

\begin{table*}[pt]
	\begin{center}
		\begin{tabular}{|c|c|c|}
			\hline
			\textbf{Name} & \textbf{Blog} & \textbf{On} \\
			\hline
			\hline
			AmarBlog & \url{www.amarblog.com} & 8$^\text{th}$ July, 2013 \\
			\hline
			Bokolom & \url{www.bokolom.com} & 10$^\text{th}$ July, 2013 \\
			\hline
			CoffeeHouserAdda & \url{www.coffeehouseradda.in} & 7$^\text{th}$ July, 2013 \\
			\hline
			CadetCollege & \url{www.cadetcollegeblog.com} & 16$^\text{th}$ July, 2013 \\
			\hline
			ChoturMatrik & \url{www.choturmatrik.net} & 7$^\text{th}$ July, 2013 \\
			\hline
			MuktoBlog & \url{www.muktoblog.net} & 9$^\text{th}$ July, 2013 \\
			\hline
			MuktoMona & \url{www.mukto-mona.com} & 10$^\text{th}$ July, 2013 \\
			\hline
			NagarikBlog & \url{www.nagorikblog.com} & 9$^\text{th}$ July, 2013 \\
			\hline
			Nirman & \url{www.nirmaaan.com} & 8$^\text{th}$ July, 2013 \\
			\hline
			Sachalayatan & \url{www.sachalayatan.com} & 8$^\text{th}$ July, 2013 \\
			\hline
			SomeWhereIn & \url{www.somewhereinblog.net} & 15$^\text{th}$ July, 2013 \\
			\hline
		\end{tabular}
		\tabcaption{Blog corpus.}
		\label{tab:blog}
	\end{center}
\end{table*}

\section{Morphological Parsing}
\label{app:morpheme}

\begin{table*}[pt]
	\begin{center}
	\begin{tabular}{|c||c|c|c|c|}
	\hline
	Corpus & Accuracy & Recall & Precision & F-Score \\
	\hline
	\hline
	Classical & 48.80\% & 40.00\% & 49.78\% & 44.38\% \\
	\hline
	Blog & 55.60\% & 48.80\% & 53.53\% & 51.05\% \\
 	\hline
	Newspaper & 54.30\% & 47.70\% & 52.68\% & 50.06\% \\
	\hline
	Merged & 56.40\% & 50.31\% & 54.00\% & 52.08\% \\
	\hline
	\end{tabular}
	\tabcaption{Performance of Undivide++ \cite{undivide} on our corpora.}
	\label{tab:morphemes}
	\end{center}
\end{table*}

Table~\ref{tab:morphemes} shows the performance of Undivide++ on our corpora.

\section{Results}
\label{app:results}

Table~\ref{tab:cpwfreq} to Table~\ref{tab:wpsfreq} show the p-values of all the
symmetric tests.  Table~\ref{tab:unichrfreq} to Table~\ref{tab:biwrdfreq} show
the p-values of all the non-symmetric tests.

% Lengths

\begin{table*}[pt]
\begin{center}
\begin{tabular}{|c||c|c|c||c|c|c|}
\hline
 & \multicolumn{3}{c||}{Blog} & \multicolumn{3}{c|}{Newspaper} \\
 \cline{2-7}
 & $\neq$ & $>$ & $<$ & $\neq$ & $>$ & $<$ \\
 \hline
 \hline
Classical & $1.66\scfm{2}$ & $8.25\scfm{1}$ & $8.33\scfm{3}$ & $6.56\scfm{4}$ & $9.14\scfm{1}$ & $3.28\scfm{4}$ \\
\hline
Blog & - & - & - & $2.09\scfm{2}$ & $9.11\scfm{1}$ & $3.28\scfm{4}$ \\
\hline
\end{tabular}
\tabcaption{K-S test results for frequency of characters per word.}
\label{tab:cpwfreq}
\end{center}
\end{table*}

\begin{table*}[pt]
\begin{center}
\begin{tabular}{|c||c|c|c||c|c|c|}
\hline
 & \multicolumn{3}{c||}{Blog} & \multicolumn{3}{c|}{Newspaper} \\
 \cline{2-7}
 & $\neq$ & $>$ & $<$ & $\neq$ & $>$ & $<$ \\
 \hline
 \hline
Classical & $8.96\scfm{2}$ & $4.43\scfm{2}$ & $2.84\scfm{2}$ & $3.03\scfm{1}$ & $9.71\scfm{1}$ & $1.52\scfm{1}$ \\
\hline
Blog & - & - & - & $2.82\scfm{3}$ & $9.74\scfm{1}$ & $1.41\scfm{3}$ \\
\hline
\end{tabular}
\tabcaption{K-S test results for frequency of syllables per word.}
\label{tab:spwfreq}
\end{center}
\end{table*}

\begin{table*}[pt]
\begin{center}
\begin{tabular}{|c||c|c|c||c|c|c|}
\hline
 & \multicolumn{3}{c||}{Blog} & \multicolumn{3}{c|}{Newspaper} \\
 \cline{2-7}
 & $\neq$ & $>$ & $<$ & $\neq$ & $>$ & $<$ \\
 \hline
 \hline
Classical & $9.79\scfm{1}$ & $8.94\scfm{1}$ & $6.41\scfm{1}$ & $9.79\scfm{1}$ & $8.94\scfm{1}$ & $6.41\scfm{1}$ \\
\hline
Blog & - & - & - & $9.99\scfm{1}$ & $8.94\scfm{1}$ & $8.94\scfm{1}$ \\
\hline
\end{tabular}
\tabcaption{K-S test results for frequency of segments (morphemes plus root word) per word.}
\label{tab:mpwfreq}
\end{center}
\end{table*}

\begin{table*}[pt]
\begin{center}
\begin{tabular}{|c||c|c|c||c|c|c|}
\hline
 & \multicolumn{3}{c||}{Blog} & \multicolumn{3}{c|}{Newspaper} \\
 \cline{2-7}
 & $\neq$ & $>$ & $<$ & $\neq$ & $>$ & $<$ \\
 \hline
 \hline
Classical & $9.97\scfm{1}$ & $7.26\scfm{1}$ & $7.26\scfm{1}$ & $9.99\scfm{1}$ & $8.35\scfm{1}$ & $8.35\scfm{1}$ \\
\hline
Blog & - & - & - & $8.64\scfm{1}$ & $4.86\scfm{1}$ & $6.06\scfm{1}$ \\
\hline
\end{tabular}
\tabcaption{K-S test results for frequency of words per sentence.}
\label{tab:wpsfreq}
\end{center}
\end{table*}

% Characters

\begin{table*}[pt]
\begin{center}
\begin{tabular}{|c||c|c|c||c|c|c||c|c|c|}
\hline
 & \multicolumn{3}{c||}{Classical} & \multicolumn{3}{c||}{Blog} & \multicolumn{3}{c|}{Newspaper} \\
 \cline{2-10}
 & $\neq$ & $>$ & $<$ & $\neq$ & $>$ & $<$ & $\neq$ & $>$ & $<$ \\
 \hline
 \hline
Classical & - & - & - & $4.63\scfm{1}$ & $2.37\scfm{1}$ & $7.93\scfm{1}$ & $0.60\scfm{1}$ & $3.09\scfm{1}$ & $6.96\scfm{1}$ \\
\hline
Blog & $9.97\scfm{1}$ & $7.30\scfm{1}$ & $7.30\scfm{1}$ & - & - & - & $7.30\scfm{1}$ & $8.38\scfm{1}$ & $7.30\scfm{1}$ \\
\hline
Newspaper & $9.67\scfm{1}$ & $6.12\scfm{1}$ & $7.30\scfm{1}$ & $9.97\scfm{1}$ & $7.30\scfm{1}$ & $8.38\scfm{1}$ & - & - & - \\
\hline
\end{tabular}
\tabcaption{K-S test results for frequency of uni-grams of characters.}
\label{tab:unichrfreq}
\end{center}
\end{table*}

\begin{table*}[pt]
\begin{center}
\begin{tabular}{|c||c|c|c||c|c|c||c|c|c|}
\hline
 & \multicolumn{3}{c||}{Classical} & \multicolumn{3}{c||}{Blog} & \multicolumn{3}{c|}{Newspaper} \\
 \cline{2-10}
 & $\neq$ & $>$ & $<$ & $\neq$ & $>$ & $<$ & $\neq$ & $>$ & $<$ \\
 \hline
 \hline
Classical & - & - & - & $6.03\scfm{2}$ & $6.39\scfm{1}$ & $3.01\scfm{2}$ & $1.52\scfm{1}$ & $7.15\scfm{1}$ & $7.64\scfm{2}$ \\
\hline
Blog & $1.32\scfm{2}$ & $2.04\scfm{2}$ & $6.60\scfm{3}$ & - & - & - & $1.32\scfm{2}$ & $2.04\scfm{1}$ & $6.60\scfm{3}$ \\
\hline
Newspaper & $1.11\scfm{11}$ & $1.0$ & $3.06\scfm{17}$ & $1.64\scfm{1}$ & $6.70\scfm{1}$ & $8.20\scfm{2}$ & - & - & - \\
\hline
\end{tabular}
\tabcaption{K-S test results for frequency of bi-grams of characters.}
\label{tab:bichrfreq}
\end{center}
\end{table*}

\begin{table*}[pt]
\begin{center}
\begin{tabular}{|c||c|c|c||c|c|c||c|c|c|}
\hline
 & \multicolumn{3}{c||}{Classical} & \multicolumn{3}{c||}{Blog} & \multicolumn{3}{c|}{Newspaper} \\
 \cline{2-10}
 & $\neq$ & $>$ & $<$ & $\neq$ & $>$ & $<$ & $\neq$ & $>$ & $<$ \\
 \hline
 \hline
Classical & - & - & - & $2.16\scfm{5}$ & $5.25\scfm{1}$ & $1.08\scfm{9}$ & $7.05\scfm{8}$ & $7.64\scfm{2}$ & $3.52\scfm{8}$ \\
\hline
Blog & $2.48\scfm{5}$ & $3.63\scfm{2}$ & $1.24\scfm{5}$ & - & - & - & $6.25\scfm{5}$ & $9.32\scfm{2}$ & $3.12\scfm{5}$ \\
\hline
Newspaper & $2.48\scfm{5}$ & $3.63\scfm{2}$ & $1.24\scfm{5}$ & $6.25\scfm{5}$ & $9.32\scfm{2}$ & $3.12\scfm{5}$ & - & - & - \\
\hline
\end{tabular}
\tabcaption{K-S test results for frequency of tri-grams of characters.}
\label{tab:trichrfreq}
\end{center}
\end{table*}

% Syllables

\begin{table*}[pt]
\begin{center}
\begin{tabular}{|c||c|c|c||c|c|c||c|c|c|}
\hline
 & \multicolumn{3}{c||}{Classical} & \multicolumn{3}{c||}{Blog} & \multicolumn{3}{c|}{Newspaper} \\
 \cline{2-10}
 & $\neq$ & $>$ & $<$ & $\neq$ & $>$ & $<$ & $\neq$ & $>$ & $<$ \\
 \hline
 \hline
Classical & - & - & - & $1.12\scfm{1}$ & $7.26\scfm{1}$ & $5.61\scfm{2}$ & $1.77\scfm{1}$ & $7.26\scfm{1}$ & $8.89\scfm{2}$ \\
\hline
Blog & $7.27\scfm{2}$ & $8.38\scfm{1}$ & $3.00\scfm{2}$ & - & - & - & $7.20\scfm{1}$ & $8.38\scfm{1}$ & $3.82\scfm{1}$ \\
\hline
Newspaper & $2.80\scfm{1}$ & $7.30\scfm{1}$ & $1.40\scfm{1}$ & $2.80\scfm{1}$ & $6.12\scfm{1}$ & $1.40\scfm{1}$ & - & - & - \\
\hline
\end{tabular}
\tabcaption{K-S test results for frequency of uni-grams of syllables.}
\label{tab:unisylfreq}
\end{center}
\end{table*}

\begin{table*}[pt]
\begin{center}
\begin{tabular}{|c||c|c|c||c|c|c||c|c|c|}
\hline
 & \multicolumn{3}{c||}{Classical} & \multicolumn{3}{c||}{Blog} & \multicolumn{3}{c|}{Newspaper} \\
 \cline{2-10}
 & $\neq$ & $>$ & $<$ & $\neq$ & $>$ & $<$ & $\neq$ & $>$ & $<$ \\
 \hline
 \hline
Classical & - & - & - & $4.44\scfm{16}$ & $1$ & $2.09\scfm{16}$ & $4.44\scfm{16}$ & $1$ & $2.09\scfm{16}$ \\
\hline
Blog & $2.22\scfm{2}$ & $5.61\scfm{2}$ & $1.11\scfm{2}$ & - & - & - & $2.22\scfm{2}$ & $1.11\scfm{2}$ & $5.61\scfm{2}$ \\
\hline
Newspaper & $9.19\scfm{8}$ & $6.06\scfm{1}$ & $4.95\scfm{8}$ & $1.98\scfm{5}$ & $1.35\scfm{1}$ & $9.92\scfm{6}$ & - & - & - \\
\hline
\end{tabular}
\tabcaption{K-S test results for frequency of bi-grams of syllables.}
\label{tab:bisylfreq}
\end{center}
\end{table*}

% Morphemes

\begin{table*}[pt]
\begin{center}
\begin{tabular}{|c||c|c|c||c|c|c||c|c|c|}
\hline
 & \multicolumn{3}{c||}{Classical} & \multicolumn{3}{c||}{Blog} & \multicolumn{3}{c|}{Newspaper} \\
 \cline{2-10}
 & $\neq$ & $>$ & $<$ & $\neq$ & $>$ & $<$ & $\neq$ & $>$ & $<$ \\
 \hline
 \hline
Classical & - & - & - & $1.4\scfm{2}$ & $6.06\scfm{1}$ & $7.31\scfm{4}$ & $2.95\scfm{4}$ & $7.26\scfm{1}$ & $1.47\scfm{4}$ \\
\hline
Blog & $6.17\scfm{3}$ & $3.08\scfm{3}$ & $6.06\scfm{1}$ & - & - & - & $8.64\scfm{1}$ & $8.35\scfm{1}$ & $4.86\scfm{1}$ \\
\hline
Newspaper & $4.44\scfm{16}$ & $2.09\scfm{16}$ & $1$ & $4.44\scfm{16}$ & $2.09\scfm{16}$ & $9.75\scfm{1}$ & - & - & - \\
\hline
\end{tabular}
\tabcaption{K-S test results for frequency of uni-grams of morphemes.}
\label{tab:unimorfreq}
\end{center}
\end{table*}

% Words

\begin{table*}[pt]
\begin{center}
\begin{tabular}{|c||c|c|c||c|c|c||c|c|c|}
\hline
 & \multicolumn{3}{c||}{Classical} & \multicolumn{3}{c||}{Blog} & \multicolumn{3}{c|}{Newspaper} \\
 \cline{2-10}
 & $\neq$ & $>$ & $<$ & $\neq$ & $>$ & $<$ & $\neq$ & $>$ & $<$ \\
 \hline
 \hline
Classical & - & - & - & $3.33\scfm{16}$ & $2.65\scfm{1}$ & $1.87\scfm{16}$ & $3.33\scfm{16}$ & $9.40\scfm{2}$ & $1.87\scfm{16}$ \\
\hline
Blog & $1.01\scfm{12}$ & $8.38\scfm{1}$ & $5.05\scfm{13}$ & - & - & - & $1.37\scfm{7}$ & $4.93\scfm{1}$ & $6.89\scfm{8}$ \\
\hline
Newspaper & $1.23\scfm{13}$ & $6.06\scfm{1}$ & $6.14\scfm{14}$ & $0$ & $1$ & $1.92\scfm{22}$ & - & - & - \\
\hline
\end{tabular}
\tabcaption{K-S test results for frequency of uni-grams of words.}
\label{tab:uniwrdfreq}
\end{center}
\end{table*}

\begin{table*}[pt]
\begin{center}
\begin{tabular}{|c||c|c|c||c|c|c||c|c|c|}
\hline
 & \multicolumn{3}{c||}{Classical} & \multicolumn{3}{c||}{Blog} & \multicolumn{3}{c|}{Newspaper} \\
 \cline{2-10}
 & $\neq$ & $>$ & $<$ & $\neq$ & $>$ & $<$ & $\neq$ & $>$ & $<$ \\
 \hline
 \hline
Classical & - & - & - & $1.37\scfm{7}$ & $3.82\scfm{1}$ & $6.92\scfm{8}$ & $7.40\scfm{11}$ & $3.82\scfm{1}$ & $3.70\scfm{11}$ \\
\hline
Blog & $4.33\scfm{8}$ & $2.04\scfm{1}$ & $2.16\scfm{8}$ & - & - & - & $7.40\scfm{11}$ & $7.30\scfm{1}$ & $3.70\scfm{11}$ \\
\hline
Newspaper & $1.01\scfm{12}$ & $7.30\scfm{1}$ & $5.05\scfm{13}$ & $2.48\scfm{5}$ & $3.82\scfm{1}$ & $1.24\scfm{5}$ & - & - & - \\
\hline
\end{tabular}
\tabcaption{K-S test results for frequency of bi-grams of words.}
\label{tab:biwrdfreq}
\end{center}
\end{table*}

\end{document}